\documentclass[runningheads]{llncs}

\usepackage[export]{adjustbox}
\usepackage[final]{eccv}

\usepackage{eccvabbrv}
\usepackage{graphicx}
\usepackage{booktabs}
\usepackage{makecell}
\usepackage{multirow}
\usepackage{enumitem}
\usepackage[accsupp]{axessibility}

\usepackage{hyperref}
\usepackage{orcidlink}

\begin{document}

\title{NumerosityVLM: A Cognitively Inspired Benchmark for Interpreting Numerosity Representations in Vision--Language Models}

\titlerunning{NumerosityVLM}

\author{
Yiming Fu\inst{1}\orcidlink{0009-0006-6165-558X} \and
Fangjun Li\inst{1} \orcidlink{0000-0002-1109-6285}\and
Xiujin Liu\inst{2} \orcidlink{0009-0000-7008-1270}\and
Ruidong Ma\inst{5}\orcidlink{0000-0002-8035-5746} \and
Hang Yu\inst{3}\orcidlink{0009-0001-4258-0192} \and
Zhichen Lu\inst{4}\orcidlink{0009-0001-8436-4519} \and
Kanwei He\inst{1}\orcidlink{0009-0001-5616-6953} \and
Alessandro Di Nuovo\inst{5}\orcidlink{0000-0003-2677-2650} \and
Angelo Cangelosi\inst{1} \orcidlink{0000-0002-4709-2243}\and
Zhegong Shangguan\inst{1}\orcidlink{0000-0002-7948-0531} \thanks{Corresponding author. Email: zhegong.shangguan@manchester.ac.uk}
}

\authorrunning{Y.~Fu et al.}

\institute{
Cognitive Robotics Lab, Department of Computer Science, The University of Manchester, Manchester M13 9PL, UK
\\
\and University of Michigan, MI 48109, USA\\
\and Tufts University, MA 02155, USA\\
\and ENSTA, Institut Polytechnique de Paris, Palaiseau 91120, France\\
\and Sheffield Hallam University, Sheffield S1 1WB, UK
}

\maketitle

\begin{abstract}
Vision-language models (VLMs) achieve strong performance on high-level multimodal tasks, yet numerosity perception, a cognitive ability that emerges in human infants before language acquisition, remains poorly understood in current models, as existing counting benchmarks entangle numerosity with correlated visual factors. We introduce a cognitively inspired diagnostic benchmark, \textbf{NumerosityVLM}, comprising 10,800 synthetic images across six controlled conditions. The benchmark orthogonally manipulates object size, spatial arrangement, and numerosity, while progressively ablating texture, shape, and color. Evaluating seven VLMs in a zero-shot setting, multi-factor analysis reveals that model architecture explains the largest proportion of performance variance (partial $\omega^{2}=0.325$), far exceeding visual conditions. Layer-wise probing further shows that linearly separable numerosity signals consistently emerge at early stages of the vision encoder, while performance differences across evaluated models are primarily associated with the language model component. Code and data are publicly available at \url{https://github.com/fuy3/NumerosityVLM-Benchmark}, and \url{https://huggingface.co/datasets/fuy3/NumerosityVLM}.

\keywords{Vision--Language Models \and Numerosity Perception \and Human-Inspired Vision \and Cognitive Benchmark \and Mechanistic Interpretability}

\end{abstract}

\section{Introduction}
\label{sec:intro}

Vision--language models (VLMs) have achieved remarkable performance on multimodal tasks such as visual question answering~\cite{sima2024drivelm}, scene understanding~\cite{simeoni2025dinov3}, and chart interpretation~\cite{xia2025chartx}, yet whether they also possess fundamental perceptual abilities remains an open question. Recent studies suggest a gap between high-level reasoning and low-level visual perception in VLMs~\cite{rahmanzadehgervi2024vision}, raising concerns that strong benchmark performance may not reflect genuine numerical understanding. Numerosity perception, which emerges in human infants before language acquisition~\cite{sarnecka2008counting}, provides a natural testbed for evaluating such abilities. Existing counting benchmarks, however, suffer from two major limitations: general-purpose benchmarks~\cite{li2023seed,fu2023mme} cover only limited numerosity ranges, whereas real-world datasets~\cite{you2023few} entangle numerosity with visual factors such as object size, density, and appearance. While cognitively inspired evaluations have been explored~\cite{testolin2025visual}, they still lack systematic control over visual variables known to influence numerosity judgments. Consequently, current benchmarks cannot reliably attribute counting failures to deficiencies in visual perception, language decoding, or reliance on shortcut cues.

To address these limitations, we introduce \textbf{NumerosityVLM}, a controlled diagnostic benchmark with six experimental conditions and evaluate seven representative open-source VLMs from four architectural families, enabling systematic investigation of numerosity perception through controlled behavioral evaluation and mechanistic analysis. Our contributions are summarized as follows:

\begin{enumerate}[label=(\roman*),nosep,leftmargin=*,topsep=2pt]
\item We introduce a cognitively inspired benchmark that orthogonally controls object size, space, and numerosity, together with progressive cue ablation, enabling systematic diagnosis of numerosity perception in VLMs.
\item We demonstrate through multi-factor attribution analysis that model architecture accounts for the largest proportion of counting performance variance, substantially exceeding visual conditions and object categories.
\item We reveal via layer-wise probing that numerosity signals consistently emerge in early vision layers, while the performance gap strongly aligns with the language model's mapping from latent representations to textual outputs.
\end{enumerate}

\section{Related Work}
\label{sec:Relate}

\paragraph{Human numerosity perception.}
Human numerosity perception relies on two complementary mechanisms: within the subitizing range ($N\le4$), enumeration is rapid and nearly error-free~\cite{railo2008role}, whereas larger numerosities are processed by the Approximate Number System (ANS), whose estimation variability follows Weber's law and increases proportionally with magnitude~\cite{rinaldi2020use,nieder2016neuronal}. Numerosity judgments are also systematically influenced by non-numerical visual cues such as object size, density, and convex hull extent~\cite{nieder2025calculating}.

\begin{figure}[t]
    \centering
    \includegraphics[width=\linewidth]{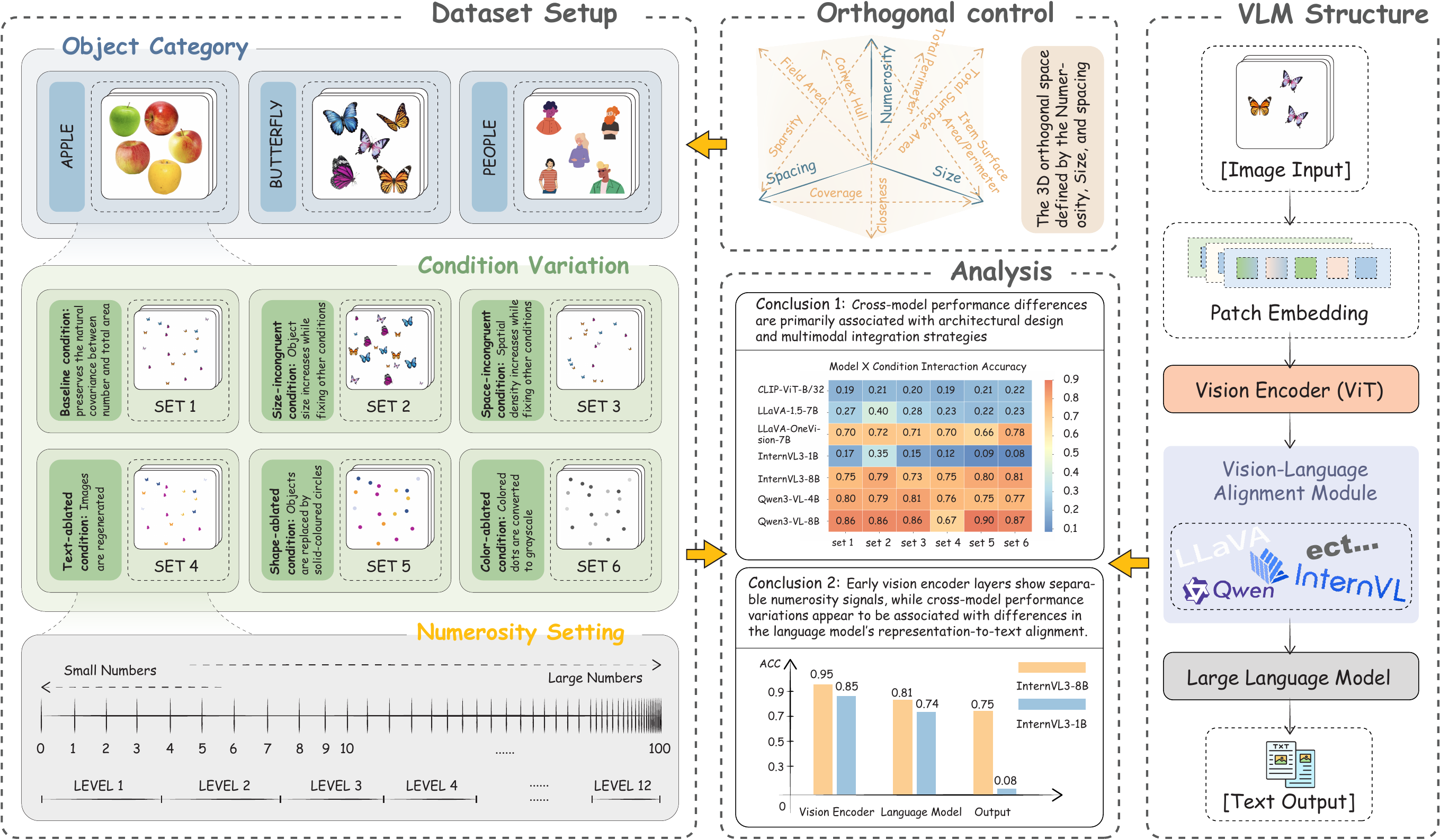}
    \caption{Overview of the proposed NumerosityVLM benchmark framework. \textbf{Left:} The diagnostic benchmark comprises three object categories, six controlled conditions, and 12 numerosity levels. \textbf{Right:} Layer-wise probing is conducted across the full VLM pipeline to localize the source of numerosity perception failures.}
    \label{fig:main}
\end{figure}

\paragraph{Counting evaluation in vision--language models.}
Object counting has become a common benchmark for evaluating VLM perception. Existing datasets are broadly categorized into general-purpose benchmarks~\cite{li2023seed,fu2023mme,sengupta2025can}, which mainly focus on small numerosities, and counting-specific datasets~\cite{nguyen2022few,ranjan2021learning,dai2024referring}, which extend to larger and denser scenes but rely on complex real-world images, making systematic error attribution difficult. Recent studies have moved beyond performance evaluation toward diagnostic analysis. CountCLIP~\cite{paiss2023teaching} and subsequent work~\cite{amini2024countgd,sengupta2025can} attributed counting failures to dataset bias, visual clutter, and attention limitations, while Testolin et al.~\cite{testolin2025visual} introduced a cognitively inspired evaluation framework. Numerosity representations have also been shown to emerge in convolutional networks~\cite{kim2021visual} and large language models~\cite{alquboj2025number}. However, existing studies still lack both systematic factor-isolated evaluation for diagnosing the visual determinants of numerosity perception and mechanistic analysis of how numerical representations are formed throughout the VLM pipeline.

\section{The Diagnostic Counting Benchmark}
\label{sec:Benchmark}
As illustrated in Figure~\ref{fig:main} and inspired by studies of human numerosity perception, we introduce NumerosityVLM, a controlled synthetic benchmark designed for the mechanistic evaluation of vision–language models (VLMs). 

\subsection{Design Principles}

All images are synthesized on a $1024 \times 1024$ pixel, uniformly white canvas, with non-overlapping objects constrained to maintain a minimum pairwise inter-object distance greater than 5 pixels and to lie entirely within the image boundaries. The dataset is generated automatically using Python scripts, and for each image we record metadata comprising individual surface area (ISA), total surface area (TSA), convex hull area, mean spatial sparsity, and object coordinates.

\begin{figure}[t]
    \centering
    \includegraphics[width=1\linewidth]{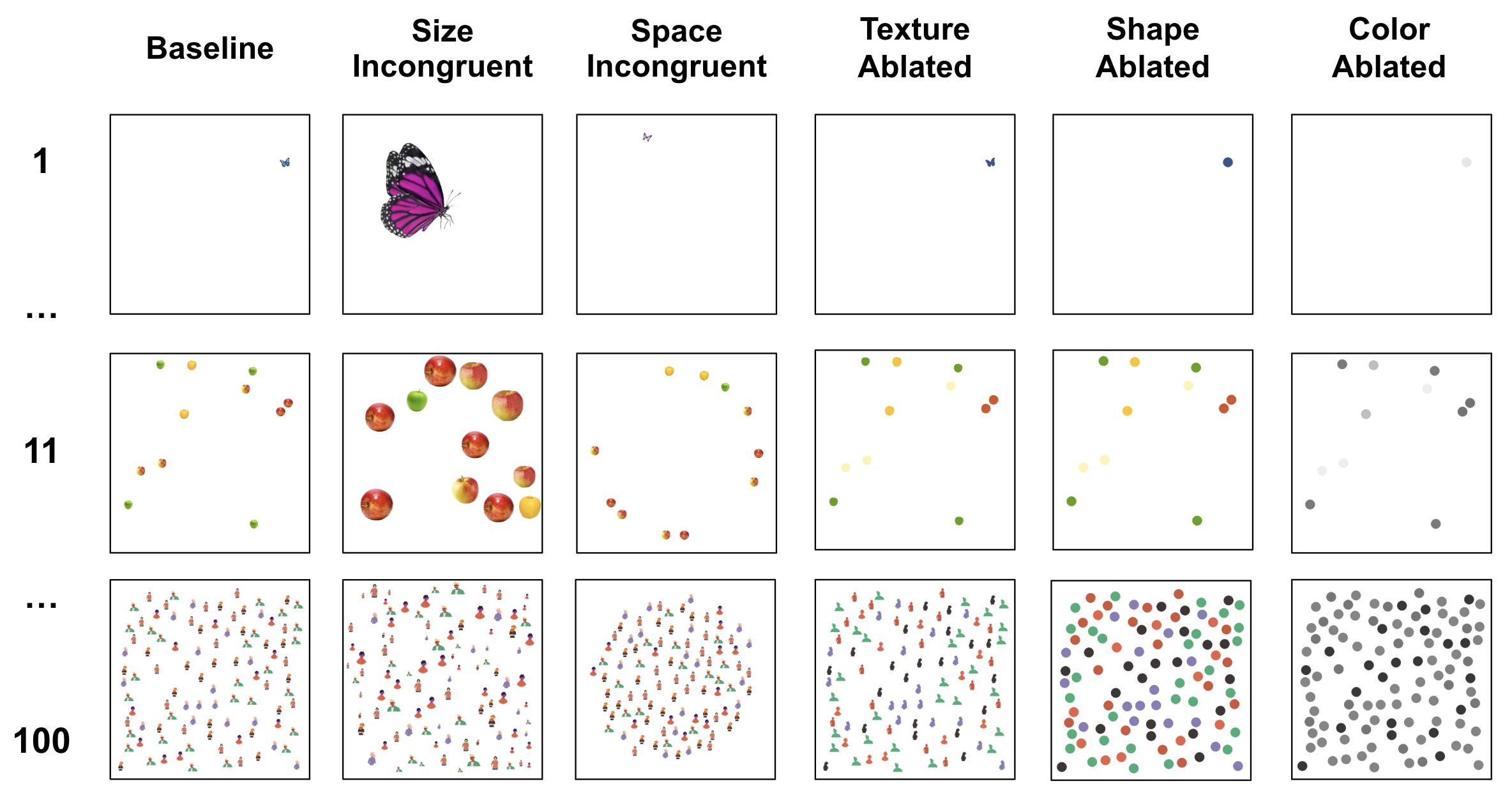}
    \caption{
    Representative dataset stimuli under the six controlled conditions. Columns show the designed controlled conditions. Rows illustrate numerosities of 1, 11, and 100 using examples from the apple, butterfly, and human-figure categories.}
    \label{fig: dataset}
\end{figure}

\subsection{Controlled Conditions}
The benchmark comprises two complementary groups of conditions: (i) three orthogonal control conditions that independently manipulate object size and spatial arrangement, and numerosity. (ii) three visual-cue ablation conditions that progressively remove texture, shape, and color information. Examples of samples from all six conditions are shown in Figure~\ref{fig: dataset}.

\subsubsection{Orthogonal Control Conditions} Let $N$ denote the number of objects, ISA the individual surface area, and TSA $= N \cdot \text{ISA}$ the total surface area.
\paragraph{Baseline.} ISA is fixed at $50 \times 50$ pixels$^2$, so TSA increases monotonically with $N$. Object positions are randomly sampled. This setting preserves the natural covariance between numerosity and total area.


\paragraph{Size-Incongruent.} TSA is fixed at $2.5\times10^5$ pixels$^2$, while ISA varies inversely with $N$, subject to $\mathrm{ISA}>28\times28$ pixels$^2$ to avoid overly small objects. This isolates numerosity from total occupied area.

\paragraph{Space-Incongruent.} ISA is fixed at $50\times50$ pixels$^2$ while objects are constrained within a convex hull of approximately fixed perimeter. As $N$ increases, spatial density increases correspondingly. This dissociates numerosity from convex hull area and introduces controlled crowding.

\subsubsection{Visual Cue Ablation Conditions}

All three ablation conditions reuse the coordinates and ISA from the Baseline condition, ensuring that positions and sizes are identical. Each condition removes one visual cue in a progressive chain.

\paragraph{Texture-Ablated.} Textured objects are replaced with simplified shape templates rendered using the dominant color of each category. This examines whether predictions depend on texture cues.

\paragraph{Shape-Ablated.} Objects are further replaced by solid-colored dots with the same ISA and color as in the Texture-Ablated condition. This removes semantic shape information while preserving spatial configuration and color statistics.

\paragraph{Color-Ablated.} Colored dots are converted to grayscale while preserving position, size, and luminance. This isolates the influence of color on numerosity perception.

\subsection{Numerical Range and Dataset Statistics}

To cover both the subitizing and approximate number system (ANS) regimes, we include all integers from 1 to 4 and logarithmically spaced values from 5 to 100 with a ratio of approximately 1.45:
\begin{equation}
\{1,2,3,4,5,7,11,18,29,46,74,100\}.
\end{equation}
Across all six conditions, the benchmark includes three object categories: apples, butterflies, and human figures. Each category contains five subtype variations that differ in visual appearance, such as color and shape, as illustrated in the \textit{Object Category }section of Figure~\ref{fig:main}. For each image, 1--5 subtypes are randomly sampled according to the target numerosity, providing within-category visual diversity. For each condition, object category, and numerosity level, we generate 50 random instances, yielding a total of $6 \times 3 \times 12 \times 50 = 10{,}800$ images.





\section{Experimental Setup}
\label{sec:experiment}


We evaluate seven representative open-source VLMs on the proposed benchmark to identify the main factors affecting numerosity perception and to assess potential biases or learning shortcuts. Both open-ended and closed-ended prompt strategies are compared. All experiments are conducted on a single NVIDIA L4 GPU (24 GB) using official pre-trained weights in BF16 precision.

\subsection{Model Selection}

We select seven open-source models spanning four VLM architectures, as detailed in Table~\ref{tab:model-summary}. CLIP~\cite{hafner2021clip} aligns visual and text embeddings via contrastive learning. LLaVA~\cite{liu2023llava} projects visual features into the LLM token space through a lightweight MLP. InternVL3~\cite{zhu2025internvl3} adopts multi-modal joint pretraining with tighter vision-language fusion. Qwen3VL~\cite{bai2025qwen3} introduces DeepStack, injecting multi-scale visual features into corresponding LLM layers. 


\begin{table}[t]
\centering
\caption{ Overview of the seven evaluated open-source vision-language models, including the backbone architectures of their vision and language components.}
\setlength{\tabcolsep}{3pt}
\begin{tabular}{llllll} 
\toprule
Model & Version & Year & Input Size & Vision Model & Language Model \\
\midrule

\multirow{1}{*}{CLIP} 
& CLIP-ViT-B/32 
& 2021 
& 224$\times$224 
& ViT-B/32 
& Transformer\\

\midrule

\multirow{2}{*}{LLaVA} 
& LLaVA-1.5-7B-hf 
& 2023 
& 336$\times$336 
& ViT-L/14 
& Vicuna-7B \\

& \makecell[l]{LLaVA-OneVision-\\Qwen2-7B-ov-hf} 
& 2024 
& 384$\times$384 
& SigLIP 
& Qwen2-7B\\

\midrule

\multirow{2}{*}{InternVL} 
& InternVL3-1B-hf 
& 2025 
& 448$\times$448 
& InternViT
& Qwen2.5-0.5B \\

& InternVL3-8B-hf 
& 2025 
& 448$\times$448 
& InternViT
& Qwen2.5-7B \\

\midrule

\multirow{2}{*}{QwenVL} 
& Qwen3VL-4B-Instruct 
& 2025 
& Dynamic 
& Qwen3-ViT 
& Qwen3-4B\\

& Qwen3VL-8B-Instruct 
& 2025 
& Dynamic 
& Qwen3-ViT 
& Qwen3-8B \\

\bottomrule
\end{tabular}

\label{tab:model-summary}
\end{table}

\subsection{Statistical Attribution Analysis} 
\subsubsection{Prompting Strategy}

Both open-ended and closed-ended prompting strategies were employed during benchmark evaluation. The prompts corresponding to each strategy are specified as follows:

\begin{itemize}
    \item \textbf{Open-ended:} 
    \texttt{How many objects are in this image?}

    \item \textbf{Closed-ended:} 
    \texttt{How many objects are in this image? Choose one}  \\
    \texttt{answer from the following options: }  \\
    \texttt{\{1, 2, 3, 4, 5, 7, 11, 18, 29, 46, 74, 100\}}
\end{itemize}

To ensure consistent evaluation, identical prompt templates are applied across all compatible models. Since CLIP does not support text generation, it is evaluated only with closed-ended prompts, while open-ended prompting is applied to the other selected generative VLMs. Numeric responses were batch-extracted from model outputs using heuristic rules and regular expressions.


\subsubsection{Evaluation Matrix}
To further evaluate the numerosity perception of vision-language models, we define five complementary metrics to quantify overall performance and error magnitude. Accuracy measures exact-match correctness between predicted and ground-truth counts. Mean Absolute Error (MAE) quantifies the average absolute deviation:
\begin{equation}
\mathrm{MAE} = \frac{1}{n} \sum_{i=1}^{n} |\hat{y}_i - y_i|,
\end{equation}
where $n$ is the total number of samples. To account for scale effects across different numerosities, we compute Normalized Absolute Error (NAE):
\begin{equation}
\mathrm{NAE} = \frac{1}{n} \sum_{i=1}^{n} \frac{|\hat{y}_i - y_i|}{y_i},
\end{equation}
which normalizes the absolute deviation by the ground-truth value. This normalization is motivated by Weber's law, reflecting the ratio-dependent nature of human numerosity perception and allowing larger deviations for higher counts without disproportionately affecting performance.

In addition, we introduce MAE$_\mathrm{error}$ and NAE$_\mathrm{error}$, which are averaged only over mispredicted samples to characterize the magnitude of errors:
\begin{equation}
\mathrm{MAE}_\mathrm{error} = \frac{1}{m} \sum_{i=1}^{m} |\hat{y}_i - y_i|, \quad
\mathrm{NAE}_\mathrm{error} = \frac{1}{m} \sum_{i=1}^{m} \frac{|\hat{y}_i - y_i|}{y_i},
\end{equation}
where $m$ denotes the number of mispredicted samples.

\subsubsection{Multi-factor Contribution Analysis}
To quantify the contributions of different factors to counting performance, we perform a multi-factor ANOVA on image-level Accuracy with three categorical factors, \textit{Model}, \textit{Condition}, and \textit{Subset}, together with their pairwise interactions. Effect sizes are quantified using partial $\omega^2$, providing a statistical attribution of performance differences.

\subsection{Systematic Bias Analysis}

\subsubsection{Visual Bias Analysis}

To evaluate whether controlled visual factors induce systematic performance biases, we conducted a series of hierarchical controlled comparisons following the dataset's factor-isolation strategy. Analyses were performed separately within two model groups stratified by overall performance. Specifically, we compared: (1) \textit{Baseline} vs.\ \textit{Size-Incongruent} and \textit{Baseline} vs.\ \textit{Space-Incongruent} to assess robustness under size and spatial controls; (2) \textit{Baseline} vs.\ \textit{Texture-Ablated} to isolate the effect of removing semantic texture cues; (3) \textit{Texture-Ablated} vs.\ \textit{Shape-Ablated} to evaluate the contribution of structured shape information relative to minimal dot stimuli; and (4) \textit{Shape-Ablated} vs.\ \textit{Color-Ablated} to measure the influence of color cues. Statistical significance was assessed using a generalized estimating equation (GEE) framework. 

\subsubsection{Numerosity Bias Analysis}
To examine magnitude-dependent counting behavior, we further computed residuals as $\text{residual}=\hat{y}-y$ and analyzed their distribution across numerosity levels. Grouped mean residuals reveal systematic over- or under-counting and potential scale-dependent bias.


\subsubsection{Layer-wise Numerosity Probing} 
To better understand the internal mechanisms underlying numerosity perception in vision-language models and to analyze potential sources of performance differences, we conduct a layer-wise numerosity probing study using the proposed strictly controlled synthetic dataset. 
Let $M$ denote a VLM under evaluation. For each layer $i$, we denote its hidden representation as
\begin{equation}
    h_i = \{ x_{i,0}, x_{i,1}, \dots, x_{i,N} \},
\end{equation}
where each $x_{i,j} \in \mathbb{R}^d$ is a token embedding in the latent space of layer $i$.

We extract representations throughout the full multimodal pipeline, including (i) the visual embedding layer, (ii) the outputs of the Attention and MLP submodules in each Vision Encoder block, (iii) each layer of the vision-language alignment module, and (iv) each decoder layer of the language model. Directly training a probe on the full token set $h_i$ is undesirable due to its high dimensionality. We therefore adopt two aggregation strategies. For the global token representation (e.g., a CLS token), we use
\begin{equation}
    h_i^{\mathrm{global}} = x_{i,0},
\end{equation}
while for the patch representations, we compute the arithmetic mean over the remaining tokens:
\begin{equation}
    h_i^{\mathrm{mean}} = \frac{1}{N} \sum_{j=1}^{N} x_{i,j}.
\end{equation}
These complementary strategies enable comparison between a dedicated global summarization token and a spatially aggregated latent representation.


For each aggregated feature, we train a linear Support Vector Machine (SVM) with a fixed regularization parameter ($C=1$) using data from the Shape-Ablated condition, while keeping the backbone model frozen. To verify that our observations are not sensitive to probe hyperparameters, we additionally perform a grid search over $C \in \{0.01, 0.1, 1, 10, 100\}$ for one representative model.  Layer-wise probing accuracy is then compared with the model's final textual counting performance to assess whether numerosity information is linearly decodable at intermediate stages and how such decodability relates to end-task performance.

\subsubsection{Language-backbone Numerical Geometry}
Motivated by evidence that LLMs may encode approximately logarithmically compressed numerical representations~\cite{alquboj2025number}, we examined whether differences in language-backbone number geometry align with differences in VLM counting performance. For each \(n\in\{1,\ldots,100\}\), we presented the prompt ``The number is \(n\).'' to each language backbone and extracted its hidden representation from a common intermediate transformer layer (\(L=12\)). We applied Principal Component Analysis (PCA) to the 100 representations and denoted the projection of each number onto the first principal component as \(s_n\). We quantified numerical ordering using the Spearman correlation between \(n\) and \(s_n\), and compared logarithmic, \(s_n=a\log(n)+b\), and linear, \(s_n=an+b\), fits using \(R^2\). These geometry measures were descriptively compared with the corresponding VLMs' end-task exact-match accuracy.


\begin{table}[t]
\centering
\caption{Overall performance comparison across different VLMs.}
\label{tab:model_acc}
\begin{tabular}{lccc}
\toprule
Model & Accuracy (\%) $\uparrow$ & MAE$_\mathrm{error}$ $\downarrow$ & NAE$_\mathrm{error}$ $\downarrow$ \\
\midrule
InternVL3-1B & 15.56 & 21.40 & 2.24 \\
CLIP-ViT-B/32 & 20.39 & 24.06 & 2.97 \\
LLaVA-1.5-7B & 27.24 & 32.23 & 1.95 \\
\midrule
LLaVA-OneVision-Qwen2-7B & 71.28 & \textbf{18.14} & 0.35 \\
InternVL3-8B & 77.21 & 18.47 & 0.35 \\
Qwen3VL-4B-Instruct & 77.73 & 24.97 & 0.36 \\
Qwen3VL-8B-Instruct & \textbf{83.97} & 20.85 & \textbf{0.31} \\
\bottomrule
\end{tabular}
\end{table}

\section{Results}
\label{sec:Result}

\subsection{Statistical Attribution Analysis}
\label{sec: model attribution}
\subsubsection{Overall Performance} 
\begin{figure}[t]
    \centering
    \includegraphics[width=\linewidth]{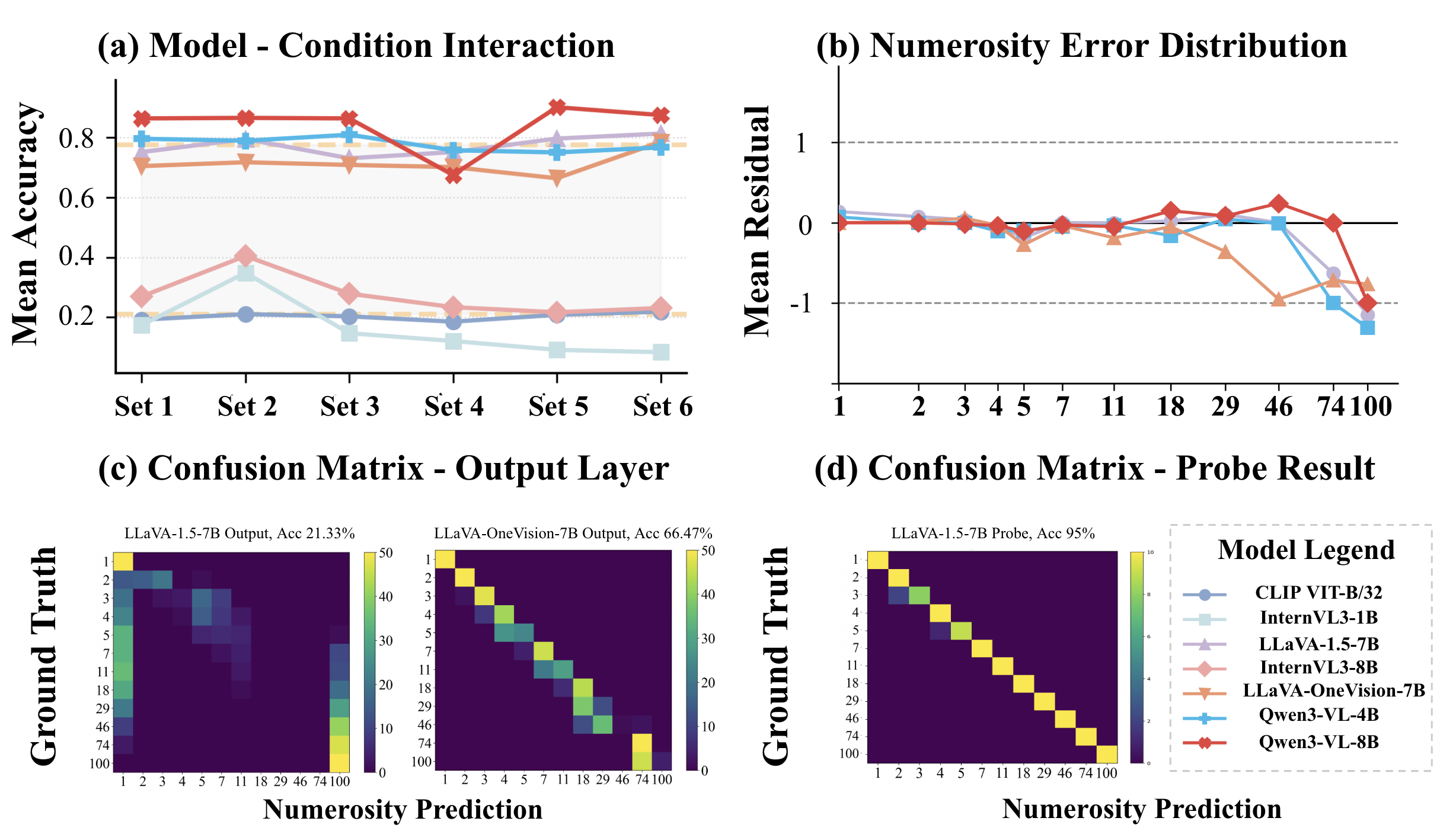}
    \caption{Comprehensive evaluation of numerosity understanding in VLMs. (a--b) Accuracy across conditions and mean residuals indicating underestimation at higher numerosity. (c--d) Output-layer and linear-probe confusion matrices for selected models.}
    \label{fig: condation}
\end{figure}

~\paragraph{Closed-Ended Setting.} As shown in Table~\ref{tab:model_acc}, the evaluated models cluster into two performance tiers. Tier-1 mainly consists of recent architectures with larger parameter scales, achieving a mean accuracy of 77.55\% $\pm$ 5.19 and a mean NAE$_{\mathrm{error}}$ of 0.34 $\pm$ 0.02. In contrast, Tier-2 models obtain a substantially lower mean accuracy of 21.06\% $\pm$ 5.87 and a higher mean NAE$_{\mathrm{error}}$ of 2.39 $\pm$ 0.53, slightly above the 8.34\% random baseline. Figure~\ref{fig: condation} (c) further reveals distinct error patterns between the two tiers: Tier-1 models exhibit more symmetric error distributions around ground-truth counts, whereas Tier-2 models fail to achieve fine-grained numerical discrimination and tend to produce a limited set of repeated outputs regardless of the true count. Detailed per-model results are provided in Appendix Table~1. To assess the benchmark's generalization beyond open-source models, we further evaluate the proprietary frontier Gemini 3 under the same protocol, with behavioral results reported in Appendix Figure~1.

\paragraph{Open-Ended Setting.} Quantitative metrics are omitted for the open-ended setting due to context-length limitations that caused some responses to be truncated before valid numerical answers could be extracted. Representative outputs and error distributions are shown in Appendix Table~2 and Appendix Figure~2.


\begin{table}[htbp]
\centering
\caption{Performance comparison in details across different dataset conditions.}
\label{tab:performance}
\begin{adjustbox}{max width=\textwidth}
\begin{tabular}{l l c c c}
\toprule
Group & Dataset Condition & Accuracy (Mean ± Std) & MAE (Mean ± Std) & NAE (Mean ± Std) \\
\midrule
Tier-1 & Baseline & 0.7780 ± 0.0586 & 21.5381 ± 1.4629 & 0.3316 ± 0.0340 \\
Tier-1 & Size-Incongruent & 0.7909 ± 0.0523 & 22.7540 ± 2.5135 & 0.3730 ± 0.0436 \\
Tier-1 & Space-Incongruent & 0.7772 ± 0.0619 & 23.2677 ± 4.0634 & 0.3493 ± 0.0422 \\
Tier-1 & Texture-Ablated & 0.7205 ± 0.0344 & 19.1333 ± 4.7798 & 0.3641 ± 0.0184 \\
Tier-1 & Shape-Ablated & 0.7771 ± 0.0853 & 18.6371 ± 4.1403 & 0.3171 ± 0.0247 \\
Tier-1 & Color-Ablated & 0.8092 ± 0.0408 & 18.2291 ± 2.9610 & 0.3501 ± 0.0203 \\
\midrule
Tier-2 & Baseline & 0.2019 ± 0.0518 & 27.2530 ± 8.9281 & 2.9104 ± 1.0483 \\
Tier-2 & Size-Incongruent & 0.3220 ± 0.0804 & 24.1742 ± 5.5354 & 1.2958 ± 0.2227 \\
Tier-2 & Space-Incongruent & 0.2087 ± 0.0541 & 25.9040 ± 5.1623 & 2.9413 ± 1.4558 \\
Tier-2 & Texture-Ablated & 0.1817 ± 0.0464 & 27.8706 ± 4.6050 & 2.6056 ± 0.3610 \\
Tier-2 & Shape-Ablated & 0.1718 ± 0.0568 & 28.3641 ± 4.4524 & 2.9368 ± 0.6242 \\
Tier-2 & Color-Ablated & 0.1783 ± 0.0669 & 21.8029 ± 2.2649 & 1.6320 ± 0.9012 \\
\bottomrule
\end{tabular}
\end{adjustbox}
\end{table}

\begin{table}[t]
\centering
\caption{Pairwise statistical comparisons across experimental conditions. Reported $p$-values are shown before and after multiple-comparison correction. Significance levels: $^{*}p<0.05$, $^{**}p<0.01$, $^{***}p<0.001$; $-$ indicates no significant difference.}
\label{tab:pairwise_tests}
\begin{tabular}{lllll}
\toprule
Group & Comparison & Raw $p$-value & Corrected $p$-value & Sig. \\
\midrule
Tier-1 & Baseline vs Size-incongruent & $1.96\times10^{-3}$ & $2.75\times10^{-3}$ & $^{**}$ \\
Tier-1 & Baseline vs Space-incongruent & $8.31\times10^{-1}$ & $8.49\times10^{-1}$ & $-$ \\
Tier-1 & Baseline vs Texture-ablated & $1.63\times10^{-29}$ & $5.72\times10^{-29}$ & $^{***}$ \\
Tier-1 & Texture-ablated vs Shape-ablated & $3.87\times10^{-40}$ & $2.71\times10^{-39}$ & $^{***}$ \\
Tier-1 & Shape-ablated vs Color-ablated & $1.79\times10^{-14}$ & $4.17\times10^{-14}$ & $^{***}$ \\
\midrule
Tier-2 & Baseline vs Size-incongruent & $2.53\times10^{-52}$ & $1.77\times10^{-51}$ & $^{***}$ \\
Tier-2 & Baseline vs Space-incongruent & $6.79\times10^{-1}$ & $6.79\times10^{-1}$ & $-$ \\
Tier-2 & Baseline vs Texture-ablated & $5.81\times10^{-12}$ & $1.36\times10^{-11}$ & $^{***}$ \\
Tier-2 & Texture-ablated vs Shape-ablated & $3.68\times10^{-2}$ & $5.15\times10^{-2}$ & $-$ \\
Tier-2 & Shape-ablated vs Color-ablated & $1.29\times10^{-1}$ & $1.51\times10^{-1}$ & $-$ \\
\bottomrule
\end{tabular}
\end{table}

\subsection{Systematic Bias Analysis}

\subsubsection{Visual Bias Analysis}

\paragraph{Orthogonal control conditions.}
Under the Size-Incongruent condition (total area fixed; numerosity inversely correlated with individual surface area), models exhibited a statistically significant improvement over Baseline ($p<0.01$ for Tier~1, $p<0.001$ for Tier~2) across both performance tiers, with Tier~2 models increasing their overall accuracy by 12.01\%. In contrast, under the Space-Incongruent condition (convex hull perimeter fixed), performance remained comparable to Baseline across both tiers: Tier~1 models achieved 77.72\% mean accuracy versus 77.80\% at Baseline, while Tier~2 models scored 20.87\% compared to 20.19\%, with no statistically significant differences observed.

\paragraph{Visual cue ablation conditions.}
Under the Texture-Ablated condition, where textured object images were replaced with abstract, shape-matched colored silhouettes, models showed a significant decline relative to Baseline ($p < 0.001$) across both tiers. Tier~1 models improved when individual visual cues were selectively removed, where accuracy increased from 72.05\% $\pm$ 3.44\% at the Texture-Ablated condition to 77.71\% $\pm$ 8.53\% under shape ablation and further to 80.92\% $\pm$ 4.08\% with color ablation ($p < 0.001$). Tier~2 models were largely unaffected, with Texture-Ablated accuracy at 18.17\%, 17.18\% under shape ablation, and 17.83\% under color ablation, as shown in Figure~\ref{fig: condation} (a), Table~\ref{tab:performance}, and Table~\ref{tab:pairwise_tests}.

\subsubsection{Contribution Attribution} A three-way ANOVA with effect size estimation indicates that \textit{Model} explains a substantial proportion of the variance (partial $\omega^2 = 0.325$), whereas \textit{Condition} (partial $\omega^2 = 0.005$) and \textit{Subset} (partial $\omega^2 = 0.001$) contribute negligibly. Although several interaction terms reach statistical significance, their effect sizes remain small, suggesting that architectural differences are the primary driver of performance variation.

\subsection{Numerosity Bias Analysis} 
Tier-1 models perform strongly within the subitizing range (1--4), achieving 94.94\% average accuracy with a mean NAE of 0.02, approaching the near error-free level characteristic of human subitizing, as in Figure~\ref{fig: condation} (b). In the lower ANS range (5--20), models maintain 85.56\% average accuracy with a mean NAE of 0.04, and residual curves remain centered around zero. As numerosity increases from 20 to 100, average accuracy decreases to 52.15\% while the mean NAE rises to around 0.17. With further increases in numerosity, models exhibit a systematic underestimation trend: all evaluated models show consistent negative residuals over 50, with stronger models tending to exhibit a later onset of this bias.

\begin{figure}[t]
    \centering
    \includegraphics[width=\linewidth]{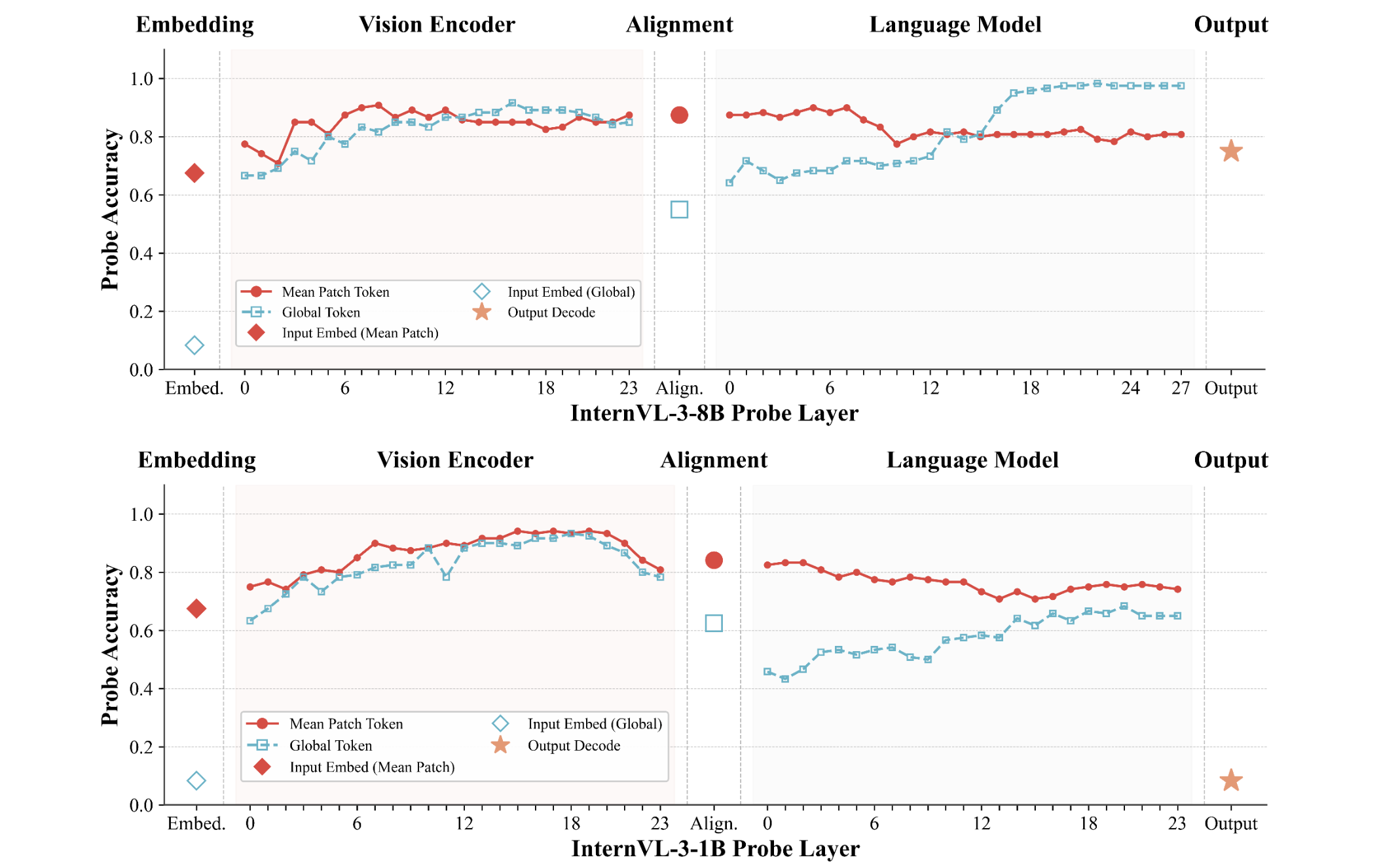}
    \caption{\textbf{Layer-wise probing results.} Numerosity signals emerge early in the vision encoder, while performance differences primarily arise during language modeling.}
    \label{fig: later-probe}
\end{figure}

\subsection{Layer-wise Analysis of Numerosity Representations}
\label{sec: pca}

Figure~\ref{fig: later-probe} presents linear probing results for representative models from the two tiers. Overall, linearly separable numerosity signals emerge in early visual layers, while cross-tier differences are primarily reflected in the language modeling and decoding stages. Similar patterns are observed across all evaluated models and under SVM hyperparameter optimization, as shown in Appendix Figures~3--7.

\textbf{Vision encoder.} Numerosity information is detectable at early patch-level representations, while global features initially remain uninformative but gradually align with patch-level signals as depth increases. By the end of the vision encoder, both tiers encode robust numerosity representations, suggesting that visual encoding is not the main source of performance variation.

\textbf{Language model.} After vision-language alignment, numerosity information becomes less stable in global representations but continues to strengthen with depth in the language model. A clear divergence emerges between tiers in later layers: Tier-1 models retain more structured numerosity information throughout decoding, whereas Tier-2 models degrade rapidly toward the output stage. Performance differences are therefore mainly associated with language-level transformation rather than visual feature extraction.

\subsection{Language-backbone Number Geometry}

\begin{figure}[t]
    \centering
    \includegraphics[width=1\linewidth]{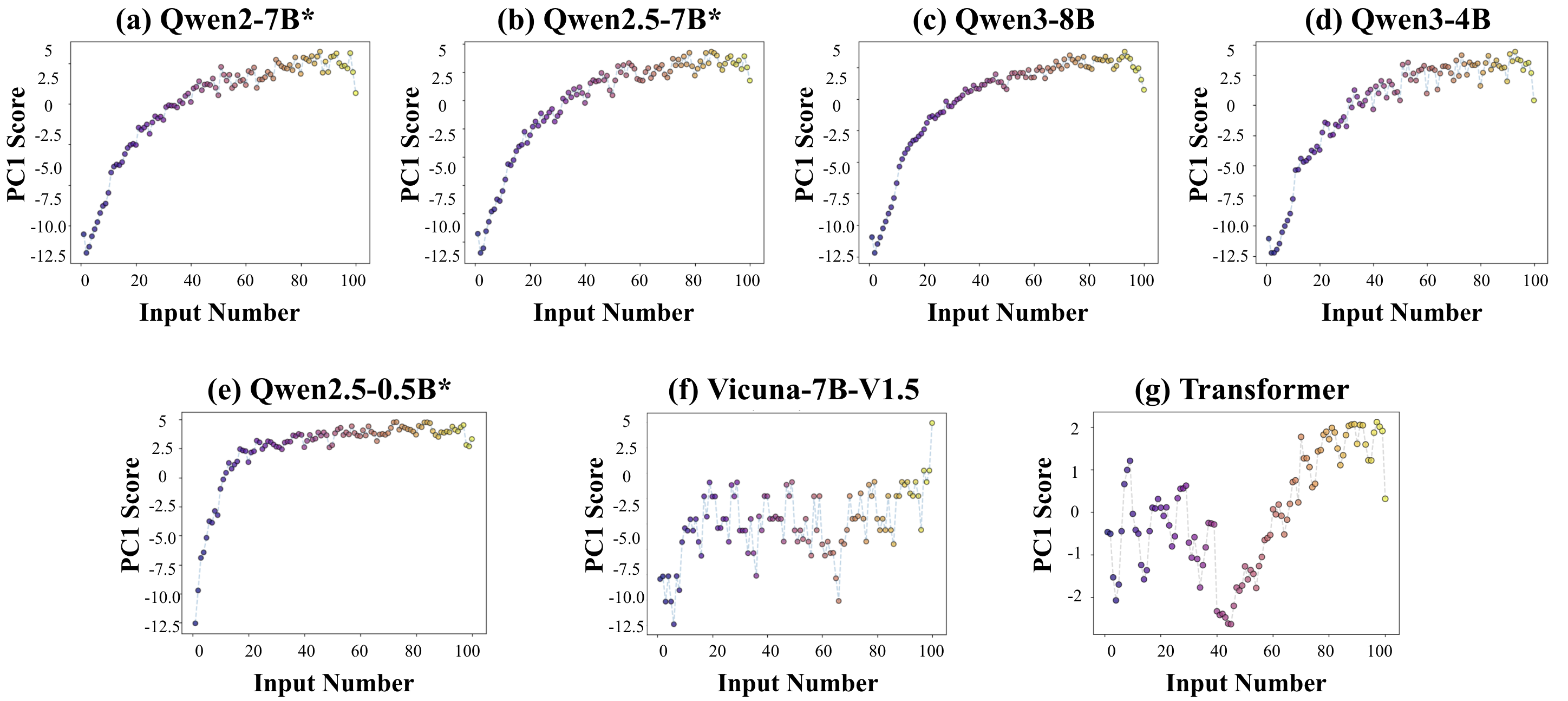}
    \caption{PC1 projections across evaluated models. Tier-1 models (a–d) show clear monotonic organization, whereas Tier-2 models (e–g) exhibit weaker or inconsistent scaling.}
    \label{fig: PCA}
\end{figure}

As shown in Figure~\ref{fig: PCA} and Table~\ref{tab:model_comparison}, Tier-1 language backbones exhibited more ordered and compressed numerical geometry, with mean Spearman $\rho=0.92\pm0.02$ and logarithmic-fit $R^2=0.95\pm0.01$, compared with $0.73\pm0.03$ for linear fits. In contrast, Tier-2 backbones showed weaker ordering ($\rho=0.64\pm0.16$) and more variable scaling, with logarithmic and linear $R^2$ values of $0.43\pm0.36$ and $0.34\pm0.12$, respectively. CLIP was the only model better characterized by a linear than a logarithmic fit ($R^2=0.37$ vs.\ $0.17$). Overall, language-backbone number geometry followed a broadly similar trend to VLM counting performance, despite imperfect correspondence across individual models.

\begin{table}[ht]
\centering
\setlength{\tabcolsep}{4pt}
\caption{Geometric and scaling analysis of internal number representations for 1--100. An asterisk (*) denotes models from the Instruct series.}

\label{tab:model_comparison}
\begin{adjustbox}{max width=\textwidth}
\begin{tabular}{l c c c c c c}
\toprule
\textbf{VLM} & \textbf{LLM} & \textbf{Spearman $\rho \uparrow$} & \textbf{PC1 Var.} & \textbf{Log $R^2 \uparrow$} & \textbf{Linear $R^2$} & \textbf{Log Slope} \\
\midrule
LLaVA-OV-7B & Qwen2-7B* & 0.9366 & 0.2714 & 0.9539 & 0.7627 & 6.5417 \\
InternVL3-8B & Qwen2.5-7B* & 0.9234 & 0.3489 & 0.9488 & 0.7332 & 9.2618 \\
Qwen3VL-8B & Qwen3-8B & 0.9211 & 0.3009 & 0.9473 & 0.7013 & 8.2357 \\
Qwen3VL-4B & Qwen3-4B & 0.8951 & 0.3867 & 0.9324 & 0.7274 & 4.5485 \\
\midrule
InternVL3-1B & Qwen2.5-0.5B* & 0.8100 & 0.4290 & 0.8439 & 0.4474 & 1.2121 \\
LLaVA-1.5-7B & Vicuna-7B-V1.5 & 0.4901 & 0.9416 & 0.2850 & 0.2104 & 3.6450 \\
CLIP-ViT-B/32 & Transformer & 0.6116 & 0.3328 & \textbf{0.1661} & \textbf{0.3715 }& 0.5987 \\
\bottomrule
\end{tabular}
\end{adjustbox}
\end{table}



\section{Conclusion and Limitations}

We introduce a cognitively inspired diagnostic benchmark, NumerosityVLM, with 10,800 strictly controlled synthetic images for evaluating numerosity perception in VLMs. Multi-factor analysis shows that model architecture explains the largest proportion of counting performance variance, while layer-wise probing reveals that numerosity representations emerge early in vision encoders and that cross-model differences primarily arise during language decoding, suggesting that the observed performance differences are more strongly associated with later representation-to-output processing than with the initial availability of visual numerosity information. The benchmark relies on synthetic stimuli with strict factor isolation, which may limit generalization to real-world scenes, and the probing results provide correlational rather than causal evidence. Future work will extend the framework to more complex real-world settings and explore causal intervention methods.

\section*{Acknowledgements}
This work was supported by ERC eTALK Project (Grant No. EP/Y029534/1) and by Innovate UK (Grant No. 10089807) through the Horizon Europe project PRIMI (Grant Agreement No. 101120727). The authors gratefully acknowledge the support of Research IT and the use of the Research VM Platform at The University of Manchester.

\bibliographystyle{splncs04}
\bibliography{main}

\end{document}